\documentclass[11pt]{article}

\usepackage[utf8]{inputenc}
\usepackage[T1]{fontenc} 
\usepackage[english]{babel}
\usepackage{amsmath,amssymb,amsthm}
\usepackage{natbib}
\usepackage{setspace}
\usepackage{tikz}
\usepackage{caption, subcaption, float}
\usepackage{array}
\usepackage{appendix}
\usepackage{longtable}
\usepackage{hyperref}

\usepackage[margin=1.2in]{geometry}

\title{Transparent and Fair Profiling in Employment Services: Evidence from Switzerland}

\author{Tim R\"az\footnote{University of Bern, Institute of Philosophy,  L\"anggassstrasse 49a, 3012 Bern, Switzerland. E-mail: tim.raez@posteo.de}}

\date{\today}

\begin{document}

\maketitle

\begin{abstract}
Long-term unemployment (LTU) is a challenge for both jobseekers and public employment services. Statistical profiling tools are increasingly used to predict LTU risk. Some profiling tools are opaque, black-box machine learning models, which raise issues of transparency and fairness. This paper investigates whether interpretable models could serve as an alternative, using administrative data from Switzerland. Traditional statistical, interpretable, and black-box models are compared in terms of predictive performance, interpretability, and fairness. It is shown that explainable boosting machines, a recent interpretable model, perform nearly as well as the best black-box models. It is also shown how model sparsity, feature smoothing, and fairness mitigation can enhance transparency and fairness with only minor losses in performance. These findings suggest that interpretable profiling provides an accountable and trustworthy alternative to black-box models without compromising performance.
\end{abstract}

\tableofcontents

\section{Introduction}

Long-term unemployment is a big social challenge, with high costs to both affected jobseekers and the public. To identify jobseekers at risk, public employment services (PES) increasingly rely on statistical profiling methods to determine jobseekers' risk of long-term unemployment (LTU). Statistical profiling promises to make profiling more accurate. Accuracy is sometimes considered to be the most important measure of successful profiling \citep{desie2019}, because it may increase the efficiency of activation measures and thus prevent LTU. 

Researchers also increasingly recognize that statistical profiling methods may create new problems and exacerbate existing ones. First, statistical profiling may provide unfair or even discriminatory predictions, putting socially salient groups at a disadvantage \citep{desie2019,allhu2020}. Second, some profiling methods that use machine learning methods are considered to be opaque (black-box), meaning that their functioning is hard to understand. Some PES nevertheless use black-box models \citep{galla2023} because they promise higher accuracy than alternatives. However, black-box models create problems for key stakeholders: They threaten to sideline PES caseworkers by automating their work. Black-box models also create an accountability gap by not providing jobseekers with a faithful explanation of their unemployment risk. They thereby prevent jobseekers from adapting their behavior, but also from recourse against predictions and decisions. Despite these problems, the use of black-box methods -- and possible alternatives -- in jobseeker profiling remains under-theorized and under-explored. The benefits of transparent models in jobseeker profiling have not been spelled out in the pertinent literature, and their accuracy has not been empirically compared to black-box models using real-life data.

This paper presents a first empirical investigation of interpretable statistical profiling. Interpretable models are machine learning models built to be inherently transparent. Both their general functioning and predictions for individual jobseekers can be understood. Using administrative data from Switzerland, this paper compares opaque and interpretable models, measuring their predictive performance, interpretability, and fairness. It provides a proof-of-concept study that interpretable models trained with administrative data can yield predictive performance only slightly worse than the best black-box models. It establishes the relevance of interpretability to stakeholders (jobseekers, caseworkers, policymakers, model developers) in profiling. It also shows that improving fairness is compatible with interpretable profiling. Interpretable profiling has the potential to increase transparency, accountability, and ultimately trust in employment services. 

The paper is structured as follows. Background on LTU profiling and related work, both from the profiling and the machine learning literature, are presented in Sec. \ref{sec:background}; the stakeholder perspective is also introduced there. Data and methods are introduced in Secs. \ref{sec:data} and \ref{sec:methods} respectively. Empirical results are given in Sec. \ref{sec:results}. A thorough contextualization of results, with a discussion of limitations and future work, are given in Sec. \ref{sec:discussion}.  The conclusion is given in Sec. \ref{sec:conclusion}.

\section{Background and Related Work}
\label{sec:background}

\subsection{Prediction of LTU}

There are three types of jobseeker profiling for LTU risk: rule-based profiling, which uses (few) demographic and administrative variables to classify jobseekers; caseworker-based profiling, which relies on the judgement of caseworkers; and statistical profiling, which uses statistical methods to predict a score or likelihood \citep{loxha2014,desie2019,desie2021}.\footnote{\citet{barne2015} additionally distinguish soft profiling, which constitutes a mixture of the other types.} Within statistical profiling, a distinction can be made between ``classical'', parametric statistical models like logistic regression, and machine learning (ML) based models. In the present paper, both classical statistical and state-of-the-art ML models will be trained and evaluated.

The evaluation of statistical and ML methods has several important dimensions. Predictive performance, the degree to which predicted risk corresponds to actual risk, is usually considered the most important metric \citep{desie2019}. The simplest measure is predictive accuracy (fraction of correct predictions). \citet{galla2023} point out that in the context of LTU profiling, predictive accuracy may be misleading, in particular if error rates are not reported.\footnote{In ML research, it is well known that accuracy can be problematic, in particular if the rate of positive and negative outcomes are imbalanced. In the case of class imbalances, it is recommended to use different metrics of predictive performance, such as AUC \citep{bradl1997}.} The importance of other performance metrics ultimately depends on the downstream use of predictions. If the consequences of false positive errors for jobseekers are moderate, misclassifications may be tolerable, but not if consequences are more severe \citep{galla2023}. In the present paper, AUC will be used to assess predictive performance, and in a (hypothetical) scenario with binary predictions, error rates will be reported in the form of a (normalized) confusion matrix (see Sec. \ref{sec:methods}).

Evaluative dimensions other than predictive performance may be equally important. First, some ML methods are black-boxes: both their inner workings and overall functioning may be hard to understand \citep{schwa2017,desie2019}. Other ML methods are interpretable and transparent by design (see Sec. \ref{sec:interp_background}). Proprietary instruments that are not publicly accessible can also be intransparent; an example is AMAS, a profiling tool used in Austria \citep{allhu2020}. Second, ML methods may lack fairness or be discriminatory by producing unequal outcomes (according to some metric) for socially salient groups such as gender, race, age, disability, etc. (see Sec. \ref{sec:fairness_background}). The main goal of the present paper is to compare methods that are both interpretable and fair to the best, state-of-the-art black-box methods.

\citet[Tab. 1]{desie2019} provide a survey of the use of statistical profiling in OECD countries, with information about the types of statistical profiling, their purposes and other properties. Of 11 countries, 8 use classical (parametric) models, three use ML or ``big data'' models. In the majority of cases, the use by both jobseekers and caseworkers is compulsory. Furthermore, most profiling tools rely on a combination of administrative data and questionnaires. The information by \citet{desie2019} is extended by \citet[Tab. 1]{galla2023}, to 13 countries, 10 of which use ``traditional'' models, while 3 use ML or ``big data'' models. Additional evaluations are reported by \citet[Tab. 1]{dossc2024}: ML methods are used in three additional countries. \citet{desie2019} also discuss problems of using ML in profiling, in particular the opacity of some ML methods. The following are key factors of successful profiling in European countries according to \citet{barne2015}. First, the appropriateness of profiling methods depends on the context, such as the downstream use of profiling like activation measures. Second, the successful use depends on the attitude and acceptance of caseworkers. 

Statistical profiling is not currently used in Swiss PES; see \citet{duell2010} for an analysis of activation measures in Switzerland. In the Swiss PES strategy for 2030, \citep{seco2030}, a tool to analyze the job market prospects of jobseekers is announced; however, it is unclear what form this tool will take. According to this strategy paper, caseworkers should focus their efforts on jobseekers with low job market prospects. Several studies have investigated the feasibility of statistical profiling in Switzerland. \citet{froel2007} tested the effectiveness of a profiling tool in 21 PES offices throughout Switzerland. In a randomized controlled trial (RCT) they found that the tool did not have a statistically significant impact on subsequent employment. They also found that the acceptance of the tool among caseworkers was low. \citet{arni2016} investigated the effectiveness of a statistical profiling tool in the Swiss canton of Fribourg. It was found that the integration of the tool into current procedures was challenging, that caseworkers evaluated the tool negatively, and that the prognostic quality of the tool was relatively low; on the other hand, an RCT found that the tool increased the speed of job market reentry.

\subsection{Interpretability}
\label{sec:interp_background}

Interpretability is the degree to which we understand aspects of ML models; it is an important research topic in ML \citep{hasti2009, biran2017}. ML models are inherently interpretable if they can be understood in virtue of their structure \citep{molna2020}. Models are globally interpretable if the entire prediction function can be understood. For black-box models, explainability methods (XAI) provide local and approximate insight into predictions \citep{adadi2018}. In a widely cited contribution, \citet{rudin2019} argues that XAI methods for black-box models are misleading because they do not faithfully mirror a model's prediction. Rudin also claims that the tradeoff between accuracy and interpretability is a myth, because interpretable models are often as accurate as black-box models. Rudin's claim about XAI is accepted. However, this paper will show that her claim about the accuracy-interpretability tradeoff is not entirely correct empirically. There is a (small) performance gap between the best black-box and the best interpretable models. \citet{lipto2016} argues that interpretability is not a well-defined notion, but has several conflicting aspects. \citet{raez2022c} agrees that there are several dimensions of interpretability, but the dimensions can be clarified for particular ML models. In the present paper, two interpretable models (LR, EBM) will be explored, and their interpretability will be spelled out (see Sec. \ref{sec:methods}).

\citet{gasse2022}, a precursor to the present paper, examined (different) interpretable models for LTU risk prediction with the same data as the present paper.\footnote{Note that the two theses \citet{gasse2022,gasse2023} have not been published and are not publicly available. However, copies can be obtained from the author of the present paper upon request via e-mail.} Gasser benchmarked the performance of three interpretable rule-ensemble methods (two versions of ruleFit and nodeHarvest). He found that the interpretable models did not significantly outperform logistic regression (LR) as an interpretable baseline, and underperformed in comparison to black-box models. The present paper extends Gasser's work by considering EBM as a further interpretable model, and finding it to perform better than the methods examined by Gasser. EBM have been applied to the prediction of social outcomes, such as academic risk prediction \citep{dsilv2023}, but not to LTU risk prediction. 

Black-box models usually provide predictions without additional information about how the prediction was arrived at. In the context of LTU risk prediction, this is problematic for jobseekers, who may have a right to obtain an explanation of how the prediction concerning them was arrived at \citep{goodm2017}. It is also problematic for caseworkers, who may find it hard to assess predictions \citep[Sec. 3.3]{delob2021}. This may decrease trust in predictions by both jobseekers and caseworkers, and it may also lead to caseworkers disengaging from a factually automated decision process, as witnessed by a Polish profiling tool \citep[Sec. 2.]{delob2021}. It is possible to understand predictions of black-box models via XAI methods to some extent. \citet{dossc2024} examine the added value of providing post-hoc explanations for a black-box model (random forest) in the context of Flemish PES. However, as pointed out above, XAI methods can be misleading. Interpretable models provide additional insight into predictions, which may help with user trust and to stem user disengagement. Caseworkers' ability to explain and justify statistical predictions to jobseekers was found to be important to them \citep{weitz2024}.

\subsection{Fairness}
\label{sec:fairness_background}

Fairness, equity, and justice have been debated for a long time. The debate on fairness in machine learning gained traction following \citet{angwi2016}. In this seminal contribution, the recidivism risk assessment instrument COMPAS was analyzed, and it was found that its prediction put Black people at a disadvantage, with a higher false positive rate and a lower false negative rate than white people. The analysis by \citet{angwi2016} was subsequently contested \citep{flore2016}; see \citet{baroc2019} for an overview.\footnote{There are now several journals and conferences dedicated, in part, to fairness in ML, for example the FAccT and AIES conferences.}

In the context of LTU profiling, fairness (or equity) has received increasing attention in recent years. \citet[Sec. 4]{kortn2023} provide an overview. \citet{desie2021} show that an ML-based tool used in Flanders (Belgium) misclassified jobseekers of foreign origin more often than jobseekers of Belgian origin, in comparison to two simple, rule-based tools. Group differences in false positives are one measure of (un-)fairness, or discrimination, in LTU profiling; this notion is adopted in the present paper. However, mitigating group differences in false positives does not constitute Fairness or Justice. For example, equalizing error rates will not correct for injustices already present in historical data.\footnote{This is another way of saying that FP is a \emph{conservative} fairness measure, which can be perfectly satisfied in the presence of historically entrenched inequalities, see \citet{raez2021}.} AMAS, a profiling tool used by Austrian Employment Services, was found to exhibit different base rates in predicted risk with respect to gender and citizenship, putting women and non-EU citizens at a disadvantage \citep{allhu2020,achte2025}. \citet{kern2021,bach2023} performed a thorough fairness audit using German administrative data, training three types of ML models. They found that different classification policies have a marked impact on outcomes of various fairness metrics for gender and nationality.

In the context of Switzerland, hiring discrimination based on race and gender is well documented \citep{hanga2021}, and hiring discrimination may translate into higher LTU risk for disadvantaged groups. It is known that women have a higher LTU rate in Switzerland, which may translate into higher predicted LTU risk \citep{zezul2024}. In these cases, unfairness is due to different base rates for groups (``ground truth'' in ML parlance) and need not necessarily manifest in different error rates. \citet{gasse2023} provides a thorough discussion of how to measure LTU risk using the same data as the present paper. Gasser considers how different fairness measures behave, and how they can be estimated statistically.

Several issues complicate unfairness and discrimination by profiling tools. First, discrimination cannot be avoided by removing sensitive attributes (gender, race etc.) from data, because other variables, like language skills or job sector, may be highly correlated with sensitive attributes \citep{desie2019, desie2021,delob2021}; such variables are called proxies. Second, if the number of sensitive attributes is high, it becomes challenging to equalize relevant inequalities, because the number of possible combinations may grow quickly. The fact that combinations of, e.g., race and gender are important is called intersectionality in the fair-ML literature \citep{zimme2022}. \citet{kern2021} investigate the intersection of gender and nationality. Third, the ultimate goal is not to equalize predictions, but the distributions of downstream utility or social goods. What is more, the LTU prediction itself may change the prediction distribution -- this is called performativity \citep{gohar2024} -- and fairness mitigation may have unintended and adverse consequences for disadvantaged groups \citep{zezul2024}.

\subsection{Stakeholders}

To clarify the policy relevance of key evaluative dimensions of LTU risk prediction -- predictive performance, fairness, interpretability --, it is useful to take a stakeholder perspective \citep{desie2019,delob2021}, and to discuss the importance of these dimensions for each stakeholder group.

\paragraph{Jobseekers} are directly affected by LTU risk predictions, and may incur benefits or costs depending on, say, access to voluntary or compulsory activation measures, and the costs associated with these measures. Jobseekers are interested in accurate predictions and not being discriminated against. They may also have the right to obtain an explanation of their risk prediction \citep{goodm2017}, either to adapt their behavior or to contest predictions for algorithmic recourse \citep{karim2022}. It could be beneficial to involve jobseekers in the use of profiling \citep{desie2019}, and this may be easier to implement with interpretable profiling. \citet{vdber2024} found that including jobseekers' self-assessment of job market reentry to predict LTU risk increased the predictive performance of a RF model.

\paragraph{Caseworkers} use LTU risk predictions and possibly further information obtained from the risk instrument to support jobseekers. They want to justify risk predictions to their clients \citep{weitz2024}. Depending on the system, they are also responsible for data collection and entry.\footnote{In a Swiss study, caseworkers criticized the increased workload created by data collection \citep{arni2016}. Data entry creates additional risks related to (inter-rater) reliability \citep{raez2024a}.} Statistical profiling may undermine the autonomy of caseworkers, in particular if profiling tools are the sole basis for decisions \citep{barne2015}. Profiling could also support and empower them if the profiling tool provides them with useful information about jobseekers' risk \citep[Sec. 3.3]{delob2021}. Whether statistical profiling is useful to caseworkers depends on the predictive performance of the tool. Additionally, caseworkers may want to keep their workload manageable, which requires that little new information needs to be gathered, and that the information provided by the profiling tool is concise. It is well known that the effectiveness of statistical profiling depends on the acceptance of the tool by caseworkers \citep{barne2015,arni2016,delob2021}.

\paragraph{Policymakers} need to make decisions about the overall prediction target and about how decisions based on predictions are used downstream, e.g., for activation measures \citep{vanla2021}. They need to weigh the utility of different outcomes, including the cost of wrong predictions (false positives, false negatives) for jobseekers as well as society at large, in order to formulate an overall objective for tool developers. Additionally, they are interested in risk predictions complying with anti-discrimination laws, and that predictions and decisions are understandable for both caseworkers and jobseekers. Finally, it can be beneficial for them to understand the impact of current policies on profiling.

\paragraph{Developers and scientists} need to be able to build, adapt, and troubleshoot the profiling tool based on prescriptions by policymakers, and based on the feedback of caseworkers and jobseekers. They need to implement the desired kinds of explanations of singular decisions. If they work for employment services, they may be interested in inspecting global model predictions. This can help them to improve data quality, to identify possible issues with data entry, and to resolve modeling issues like overfitting. Independent researchers may be interested in evaluating, auditing, and reproducibility. They need access to data as well as to models and other methods used to build the risk tool.

\paragraph{}Below we will return to these stakeholder groups and discuss to what extent EBM, the interpretable model examined in detail in the present paper, has the potential to help with key desiderata of each group.

\section{Data}
\label{sec:data}

\subsection{Description}

The clean dataset, used for training, validation, and testing, was compiled from six different raw datasets. The six raw datasets come from systems in Swiss PES, and contain data on: 1. The episode of unemployment and the unemployed person; 2. Previous and desired jobs; 3. The desired mode of work; 4. Insurance payments; 5. Job search region; 6. Outcome. A more thorough description of raw datasets is given in the appendix. The clean dataset is based on full administrative records on unemployment recipients (jobseekers) from the years 2014-2019. The dataset has 57 features, 27 of which are categorical, 30 are numerical.\footnote{After constructing dummies for categorical variables, the clean dataset contains 121 features.} No variables with aggregate economic indicators or survey data were used. For a full list of features, including a description of feature semantics, data sources, and range of values, see the appendix. The outcome, LTU, is defined as: 

\begin{itemize}

\item[] \textbf{Definition LTU:} A jobseeker is long-term unemployed (LTU), and therefore has outcome $1$, if the jobseeker receives unemployment benefits during each of the first 12 months of unemployment, otherwise they have outcome $0$.\footnote{Information on the outcome based on \citet{gasse2022}; the notion of LTU used here corresponds to Gasser's LTU$_1$.}

\end{itemize}

Each jobseeker counts once per eligibility period of two years, at the point where the jobseeker may or may not enter LTU. In the train-validate data (years 2014-2018), 20\% have outcome $1$, that is, enter LTU. In the train-test data (years 2018-2019), 18.3\% have outcome $1$. This means that there is distribution shift, because the operationalization of LTU is constant over time. The number of observations of train-validate and test set by years is given in Table \ref{tab:no_observations}.

\begin{table}[ht]
\centering
\begin{tabular}{|c|cccccc|}
\hline
Year & 2014 & 2015 & 2016 & 2017 & 2018 & 2019 \\
\hline
\# Obs.  &  164,001  & 164,945  & 178,068 & 180,799  & 175,873 &   166,065   \\
\hline
\end{tabular}
\caption{Number of observations per year.}
\label{tab:no_observations}
\end{table}

\subsection{Data Split, Cross Validation}

The clean data was split into two parts. Data from the years 2014-2018 was used to train and validate models. Data from the year 2019 was used as the test set to evaluate predictive performance. To avoid data leakage \citep{kaufm2012}, the test set was not used for training or hyperparameter tuning. Testing was performed after all other investigations, including interpretability and fairness, had been completed. The train-validate set was used for cross-validation (CV) in a moving window (or sliding window) configuration \citep{cerqu2020}. Specifically, the following split was used:

\begin{table}[ht]
\centering
\begin{tabular}{|c|cccccc|}
\hline
Fold & 2014 & 2015 & 2016 & 2017 & 2018 & 2019 \\
\hline
1     & Train    & Validate &          &          &	&       \\
2     &          & Train    & Validate &          & &       \\
3     &          &          & Train    & Validate & &       \\
4     &          &          &          & Train    & Validate & \\
\hline
5*  &    &          &          &          & Train    & Test \\
\hline
\end{tabular}
\caption{Moving window cross-validation splits. Years 2014-2018, folds 1-4: train-validate. Year 2019, fold $5^*$: test.}
\label{tab:tscv_moving_window}
\end{table}

The rationale for the time series CV split is to validate models in a realistic and ``causal'' scenario, in which predictions are only made on the basis of earlier data, as opposed to, say, a traditional 5-fold CV split, in which each of the years 2014-2018 would be set apart as the test set. The moving window split was used, first, because it reveals tendencies of distribution shift. Second, it is computationally less demanding than the growing window split, in which years up to $t-1$ are used to train and year $t$ is used to validate.

\section{Methods}
\label{sec:methods}

\subsection{Predictive Performance Metrics}

The metric used to assess the predictive performance of ML models is ROC-AUC (Area-Under-Curve of the Receiver Operating Characteristic), AUC for short \citep{fawce2006}. AUC measures the quality of predicted scores instead of (binary) decisions, like predictive accuracy. AUC takes values in $[0, 1]$, where $0.5$ corresponds to a random classifier, and $1$ to a perfect classifier. The higher the AUC is above $0.5$, the better is the classifier. Operationally, AUC is the probability that a randomly chosen positive case has a higher predicted score than a randomly chosen negative case.\footnote{The reason for using AUC instead of predictive accuracy is that AUC has several beneficial properties. First, it is a threshold-independent measure and only depends on a model's ability to order inputs by risk. Second, it is a better measure of performance under class imbalance, as in the case of the present data, with approximately 20\% of cases in the positive class.} For specific applications, a threshold can be chosen based on contextual factors, turning the predicted score into a binary risk prediction. If so, accuracy and other functions of the confusion matrix, such as false positive rate, can be used to assess predictive performance.

\subsection{Black-Box Models}

Three so-called black-box models were tested. All three are considered to be black-box models because, while they are based on decision trees, which are considered to be interpretable \citep{hasti2009, raez2022c}, they use large tree ensembles, which makes them hard to understand. For hyperparameters of all models see the appendix.

\paragraph{Random Forests (RF):} This is a tree ensemble method, based on bootstrap aggregation (bagging) and developed by \citet{breim2001}. RF is relatively simple to train and tune, and therefore a popular method \citep[p. 587]{hasti2009}. Several agencies in OECD countries have used RF for statistical profiling \citep[Tab. 1]{desie2019}.

\paragraph{Gradient Boosting (GB):} This tree method was originally developed by \citet{fried2001}. GB is more prone to overfitting than RF. According to \citet[Tab. 1]{desie2019} New Zealand has used GB for statistical profiling.

\paragraph{Extreme Gradient Boosting (XGB):} This is a modification of GB, developed by \citet{chen2016}, to make the method suitable for large datasets. It is much faster than GB and considered to be the state-of-the-art model for tabular data, see \citet{shwar2022}. If the main goal is high predictive performance, XGB may be the best choice among black-box models.

\subsection{Interpretable Models}

Two interpretable models were tested. Both of these models are globally interpretable, meaning that the entire prediction function can be visually inspected and understood \citep{molna2020, raez2022c}.

\paragraph{Logistic Regression (LR):} A linear model for classification tasks, see \citet{hasti2009}. It is interpretable via its coefficients. Its interpretability was enhanced by using the Lasso (Ibid.), also called $L_1$ regularization. By increasing $L_1$ regularization, one forces more and more coefficients to be zero, making the model smaller and thus more interpretable. Data was preprocessed before training, as required for consistent use of $L_1$ regularization. LR is the most popular LTU risk model in the OECD, with six countries using it (see survey in \citealt{desie2019}).

\paragraph{Explainable Boosting Machines (EBM):} This is a generalized additive model (GAM): the prediction function is a sum of (nonlinear) functions of individual variables (main effects) and variable pairs (interactions). It was introduced by \citet{lou2013}; the implementation by \citet{nori2019} was used.\footnote{EBM is a relatively recent model based on GAM. GAM was invented in the 1980s, see \citep{hasti1990}, and is an active area of research. It would be worthwhile to explore other variants, such as classical GAM with smoothing splines (Ibid.), sparse GAM \citep{ravik2009}, or more recent GAM based on neural networks \citep{agarw2021a}.} Internally, EBM employs gradient boosting for individual features. As a GAM, it is globally interpretable because while the functions of variables (or pairs of variables for interactions) are nonlinear, they can be visually inspected, understood, and modified, see \citet{raez2022c}. Local explanations, which provide insight into how individual predictions are made, can be constructed in the form of feature importance plots.\footnote{A (fictitious) example of a local explanation is given in the appendix (Fig. \ref{fig:local_explanation}).}

\subsection{Interpretability Enhancement for EBM and LR}

In addition to the off-the-shelf version of EBM (and LR), additional steps were taken to improve two dimensions of interpretability. The below description is more detailed because these methods are not part of the EBM implementation.

\paragraph{Sparsity:} The interpretability of both LR and EBM may benefit from sparsity, that is, from only using a subset of variables. There are different methods to achieve sparsity in linear (and additive) models. First, the Lasso, or $L_1$ regularization, enforces features to become zero ``all at once''. In the present paper, the Lasso was tested for both LR and EBM. For LR, it worked as intended; for EBM, the Lasso did not yield satisfactory sparsity while preserving predictive performance. Therefore, backward selection \citep[Sec. 3.3.]{hasti2009} was used to create sparse EBMs. In a first experiment, a full EBM with 57 main features, fit to the first training fold (year 2014), was taken as a starting point. The least important features (average feature importance) of this EBM were removed in steps of five, yielding a sequence of $(55, 50, 45, ..., 5)$ features. New EBMs were fit to these subsets, and the predictive performance was assessed. In a second experiment, three models of sparsities 45, 30, and 15 were fit to the four train-validate folds (years 2014-18), based on the 45, 30, 15 most important main features of the full EBM trained on that fold. Backward selection is a relatively rough method, but it is computationally tractable and yields a first indication of the extent to which sparsity in EBM can be achieved without large performance losses.

\paragraph{Smoothing Numerical Features:} EBM in the off-the-shelf version offer in-principle interpretability of individual feature functions by visual inspection. However, there is no guarantee that these functions are ``simple''. The predictor functions of five numerical features show large local fluctuations and quasi-discontinuities; see Figure \ref{fig:vers_verd_raw_smooth} for an example. These fluctuations were classified as artifacts by a domain expert and may be due to overfitting. Large local fluctuations are problematic for at least two reasons. First, they prevent global interpretability in that they make it hard to grasp the behavior of feature functions. Second, they are problematic in view of individual predictions and their explanation. If a jobseeker is located at a point of a feature function with a large deviation from the local trend, and this deviation is due to overfitting, it is hard to justify the corresponding risk prediction as fair.\footnote{This issue is closely related to the notion of individual fairness, according to which people with similar properties should be treated similarly \citep{dwork2012}. Thus, this is an instance where considerations of interpretability and fairness interact.} 

After testing several methods that remove artifacts in feature functions, it was decided to use smoothing cubic splines \citep[Sec. 2.8.1.]{hasti2009}. Smoothing splines are a statistical method to construct a smooth version of a ``rough'' function. In a first step, smoothing was applied to individual, off-the-shelf numerical feature functions of one sparse EBM-30 in the first fold. The EBM-30 was chosen to get an idea of how smoothing and sparsity interact. Different smoothing parameters for different features were chosen based on a tradeoff between the degree of smoothing and overall performance (AUC). In a second step, the goal was to understand how smoothing affects performance. To do so, the same degree of smoothing was applied uniformly to four sparse EBM from the four train-validate folds. Then the predictive performance in the four folds was measured.\footnote{Note that GAMs were initially trained using smoothing splines, see \citep{hasti1990}. A second method, GAMChanger, introduced by \citet{wang2021gam}, was also tested. GAMChanger allows users to manually edit single feature functions in a graphical user interface and to check how edits affect performance metrics on a data sample. Unfortunately, the functionality of GAMChanger was limited in practice -- storing results did not work as intended.}

\subsection{Fairness}

There have been many proposals to mitigate group disparities or unfairness in the context of ML. Technically, there are three types of fairness mitigation: Pre-processing (changing inputs), in-processing (changing the training process, for example using regularization), and post-processing (changing outputs of a trained model), see \citet{baroc2019}. In this paper, a post-processing method \citep{hardt2016} will be used to equalize false positives across three age groups. False positives were previously investigated in the context of LTU profiling by \citet{desie2021}. Here, a full EBM from the first fold was taken as a base model. Post-processing only affects the choice of decision thresholds and does not depend on the internal structure of the model. Therefore, the interpretability of models is preserved under post-processing. The experiment takes the following steps:

\begin{enumerate}

\item A  threshold is chosen, which yields binary risk predictions. The choice of threshold depends on the purpose for which the prediction is used. Here the threshold is chosen such that 80\% of all true positives are identified. This corresponds to a true positive rate (TPR), or recall, of $0.8$. This choice can be motivated by a requirement to make targeting efficient (see \citealt{gasse2022}). The fairness of the resulting (binary) predictor with respect to age is then examined. Specifically, the confusion matrix\footnote{The confusion matrix tabulates four statistics of binary predictors: true positives, false positives, true negatives, false negatives. The confusion matrix is normalized by the total count (TP + FP + TN + FN) to make results comparable across groups.} of three age groups (ages 15-29, 30-44, 45-65) of the resulting predictor is examined.

\item The post-processing method of \citet{hardt2016} is used to equalize false positive rates (FPR) for the three groups under an accuracy constraint. The implementation of \citet{weerts2023} was employed. The post-processing finds separate thresholds for the protected attribute (here: three age groups) such that false positive rates are equalized. In order to preserve the original goal of a given TPR to some extent, balanced accuracy is used as a constraint.\footnote{Balanced accuracy is defined as (TPR + TNR)/2; note that TNR = 1 - FPR} It is measured how post-processing affects the confusion matrix and performance metrics like accuracy.

\end{enumerate}

\section{Results}
\label{sec:results}

\subsection{Predictive Performance}
\label{sec:pred_perf}

\begin{figure}[H] 
\centering
\includegraphics[width=0.9\textwidth]{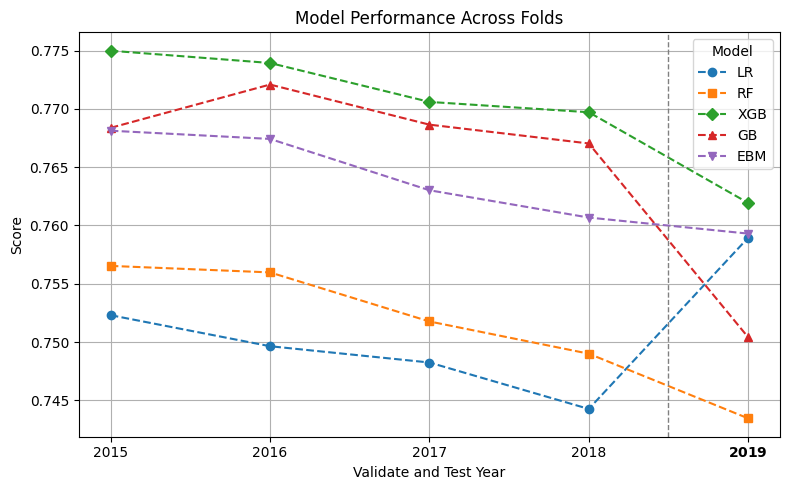}
\caption{Performance Score (AUC) for five folds. Years 2015-2018: validate; year 2019: test.}
\label{fig:scores_train_test}
\end{figure}

The results of the predictive performance (AUC) of the five models tested can be seen in Figure \ref{fig:scores_train_test}; for numbers see appendix. Two black-box models (GB, XGB) show the best overall predictive performance, with the exception of the drop of GB in the test year. XGB consistently shows the best overall performance. The relatively low performance of RF, another popular black-box model, is more surprising. Turning to the interpretable models, LR shows the overall worst relative performance of all models in validation, but improves in the test year, surpassing RF and GB. Finally, EBM results support claims in the literature that EBM performs similar to the best black-box models. EBM performs worse than the best black-box model, but the performance difference is relatively minor. \emph{The comparatively high performance of EBM is one of the main results of this paper: Interpretable models show good predictive performance in a realistic scenario while being transparent.} The performance obtained here is in a similar range as other recent empirical studies, roughly in an AUC range of 0.7-0.8 \citep{desie2019,dossc2024}. Note that the results obtained by \citet{gasse2022} on the same data show a qualitative agreement in performance with results obtained here, but AUC found by Gasser was slightly higher.\footnote{Note that while the computational cost of model fitting was not evaluated systematically, XGB was found to be fast, taking few minutes to fit four models to four folds. EBM was found to be the slowest model, taking approximately 20 minutes to fit four models to four folds. All experiments were performed on a standard desktop computer.}

The temporal evolution of predictive performances is noteworthy. The moving window time series split keeps conditions (size and temporal relation of train and validate sets) constant, which suggests that the drop in predictive performance over time is due to a shift in the data distribution. More experiments are necessary to better understand this phenomenon, for instance, to examine the effect of also including less recent data in training. The origin of the qualitative shift in the test year (drop of GB, rise of LR) is unknown; care was taken to use identical data pipelines for validation and testing. The drop of GB could be due to overfitting in validation; the reason for the rise of LR is unknown. The changes in ``deployment-like'' conditions (test year) show the importance of continuous monitoring and retraining of operational systems.

\subsection{Interpretability: Sparsity}
\label{sec:res_sparsity}

For LR, sparsity induced by $L_1$ regularization shows that using a parameter setting of $0.1$ leads to no appreciable drop in AUC with a small increase in sparsity: $115$ of $120$ features are non-zero (see Figure \ref{fig:lr_auc_sparsity}, appendix). A setting of $0.01$ leads to a small drop in AUC while $79$ of $120$ features are non-zero. The first setting, as the more conservative choice, was used in the overall performance evaluation. Dropping a third of all features may lead to a small drop in AUC for LR, which may be worth considering in terms of interpretability.

\begin{figure}[H] 
\centering
\includegraphics[width=0.9\textwidth]{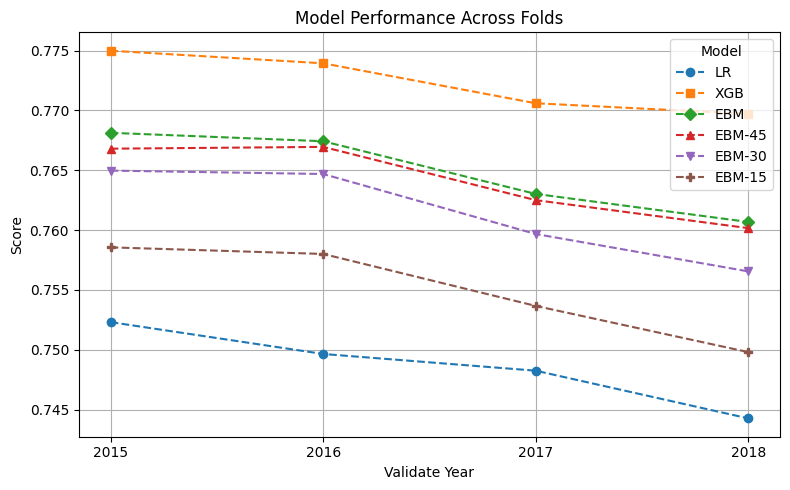}
\caption{Performance Score (AUC) for XGB, full EBM, LR, as well as three sparse EBMs with 45, 30, 15 most important main features. Results for LR, XGB, and EBM are the same as in Figure \ref{fig:scores_train_test} and included for reference.}
\label{fig:ebm_auc_sparsity_15}
\end{figure}

For EBM, the first experiment with backward selection (see Figure \ref{fig:ebm_auc_sparsity_5}, appendix), in which main features were dropped in steps of $5$, shows that dropping up to a third of main features yields only small drops in AUC, and that a model with as little as $5$ main features has an AUC of just below $0.74$, which shows that the bulk of predictive performance is due to relatively few main features. Note that the size of models drops linearly, because for each sparsity level $n$, $0.5\cdot n$ interaction terms were included. In the second experiment, the findings of the first experiment for EBM were confirmed to be robust over all validation folds, with backward selection dropping $15$ main features at a time; see Figure \ref{fig:ebm_auc_sparsity_15}. The results show that using $45$ main features only yields a relatively small drop in AUC, while the drop in performance is larger for the two sparser models. EBM-15 shows higher AUC than LR on the validation fold; this result may not hold for the test fold. Overall, these results confirm the well-known accuracy-interpretability tradeoff: The smaller and thus simpler the model, the less performant it is. They also confirm that a sizable portion of features can be dropped without large losses in predictive performance.

\subsection{Interpretability: Smoothing Numerical Features}
\label{sec:res_smooth}

The need for smoothing numerical feature functions can be motivated by looking at an example of a raw feature (sparse EBM-30), together with the smoothed version, shown in Figure \ref{fig:vers_verd_raw_smooth} (see the appendix for other smoothed features). If one examines the raw main feature of insured income, the overall trend of the plot is relatively easy to grasp, but local fluctuations are hard to understand and justify. Local fluctuations, like the large drop and spike to the very left, are likely artifacts and due to overfitting. The smoothing spline retains the overall trend of the raw function while removing artifacts. For example, the smoothing retains a clear non-linearity, with a rise in LTU risk for incomes below 4k, a dropping risk for incomes between roughly 4k and 6k, and a rise for incomes above 6k. 

\begin{figure}[H] 
\centering
\includegraphics[width=0.9\textwidth]{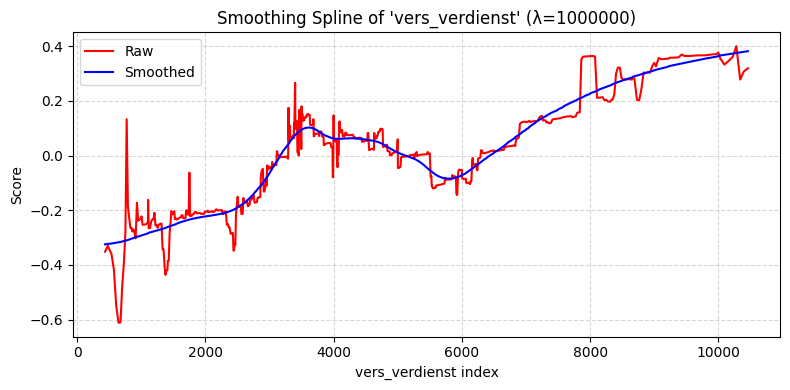}
\caption{Numerical feature `insured income' (vers\_verdienst), in raw version (off-the-shelf EBM) and after applying smoothing spline with smoothing parameter $\lambda$; insured income in CHF. The score is the contribution of the feature to the risk of LTU.}
\label{fig:vers_verd_raw_smooth}
\end{figure}

\begin{figure}[H] 
\centering
\includegraphics[width=0.9\textwidth]{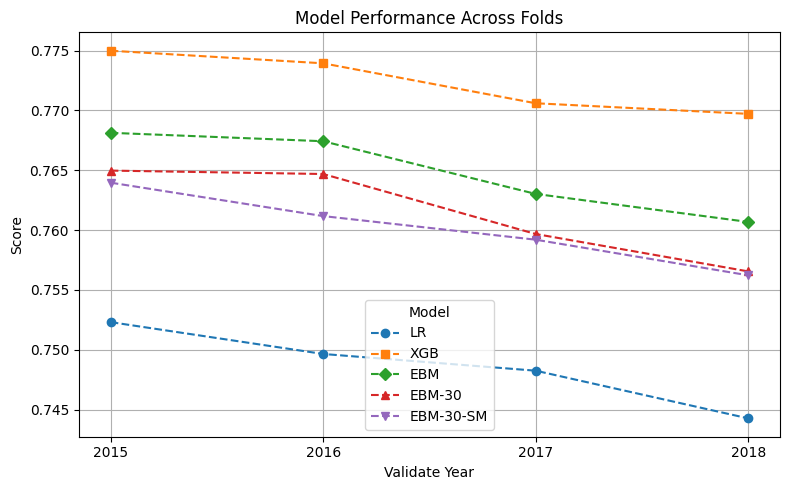}
\caption{Performance Score (AUC) for original EBM-30 (sparse EBM with 30 most important main features), and smoothed version EBM-30-SM (EBM-30 with smoothed numerical features). XGB, full EBM, LR for reference.}
\label{fig:ebm_30_sm_auc}
\end{figure}

Figure \ref{fig:ebm_30_sm_auc} shows the results of applying uniform smoothing to four sparse EBM-30 in the four train-validate folds and the measuring performance.\footnote{In each of the four folds, smoothing was applied to the following features: insured income, age, rate of desired employment, \# months of previous contributions (see Fig. \ref{fig:vers_verd_raw_smooth} and appendix). A further feature, rate of previous employment, was also smoothed, but not further investigated because it was not in the top 30 main features in all four folds.} There is a relatively modest overall drop in predictive performance from EBM-30 to the smoothed version EBM-30-SM. This is, again, a manifestation of the accuracy-interpretability tradeoff. The drop is more appreciable in the $2016$ fold, which shows that some performance risk is incurred by smoothing. It should be stressed that there is a lot of opportunity for tuning the smoothing parameter based on various considerations. For example, smoothing can be used more aggressively for features with lower average importance, and more conservatively for features with higher importance.

\subsection{Fairness}

\begin{figure}[H] 
     \centering
     \begin{subfigure}{0.49\textwidth}
         \centering
         \includegraphics[width=\textwidth]{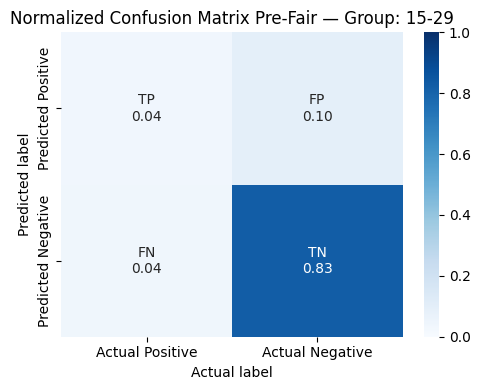}
         \caption{Age Group 15-29}
         \label{fig: ...}
     \end{subfigure}
     \hfill
     \begin{subfigure}{0.49\textwidth}
         \centering
         \includegraphics[width=\textwidth]{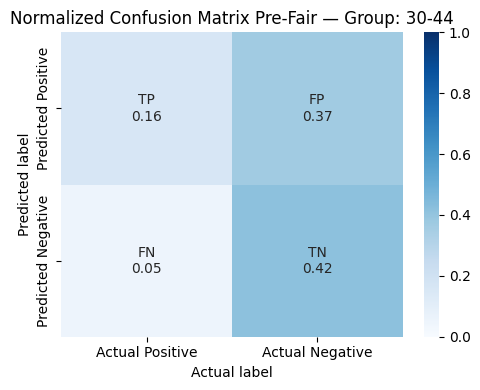}
         \caption{Age Group 30-44}
         \label{fig: ...}
     \end{subfigure}
       \bigskip
     \begin{subfigure}{0.49\textwidth}
         \centering
         \includegraphics[width=\textwidth]{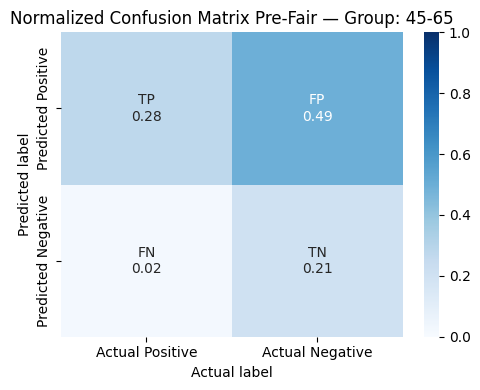}
         \caption{Age Group 45-65}
         \label{fig: ...}
     \end{subfigure}
        \hfill
             \begin{subfigure}{0.49\textwidth}
         \centering
         \includegraphics[width=\textwidth]{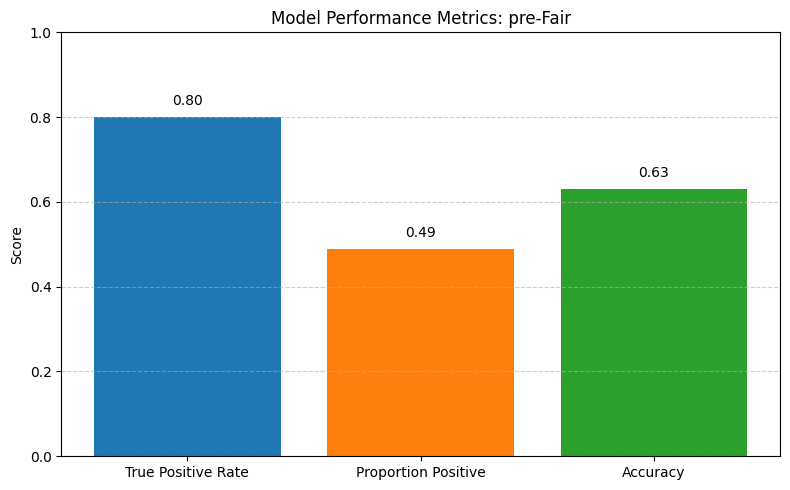}
         \caption{Performance Metrics}
         \label{fig: ...}
     \end{subfigure}
        \caption{Confusion matrices (normalized) for three age groups, and overall performance metrics, before fairness mitigation.}
        \label{fig:confusion_matrices_prefair}
\end{figure}

Figure \ref{fig:confusion_matrices_prefair} shows the results of choosing the threshold at which 80\% of all true positives are identified (TPR, see performance metrics), making the prediction binary. It can be observed that (normalized) false positives (FP) vary quite strongly between the three age groups: normalized false positives go up with age. This is to be expected, as the LTU risk also rises with age. Note that the accuracy is moderate at 63\%, but that accuracy is not the objective; the threshold was chosen for the given TPR.

\begin{figure}[H] 
     \centering
     \begin{subfigure}{0.49\textwidth}
         \centering
         \includegraphics[width=\textwidth]{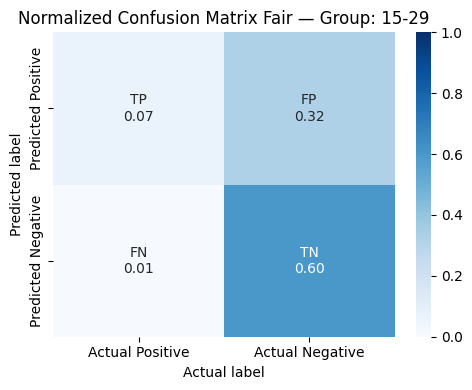}
         \caption{Age Group 15-29}
         \label{fig: ...}
     \end{subfigure}
     \hfill
     \begin{subfigure}{0.49\textwidth}
         \centering
         \includegraphics[width=\textwidth]{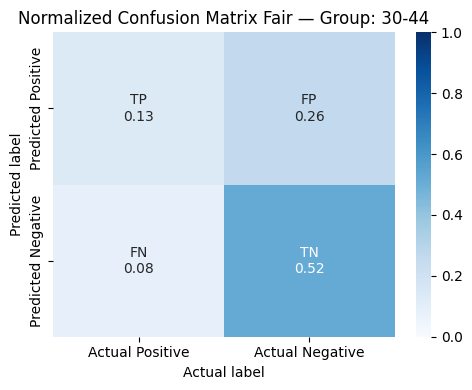}
         \caption{Age Group 30-44}
         \label{fig: ...}
     \end{subfigure}
       \bigskip  
         
     \begin{subfigure}{0.49\textwidth}
         \centering
         \includegraphics[width=\textwidth]{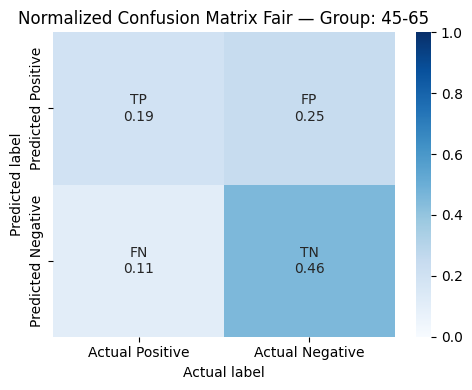}
         \caption{Age Group 45-65}
         \label{fig: ...}
     \end{subfigure}
        \hfill
             \begin{subfigure}{0.49\textwidth}
         \centering
         \includegraphics[width=\textwidth]{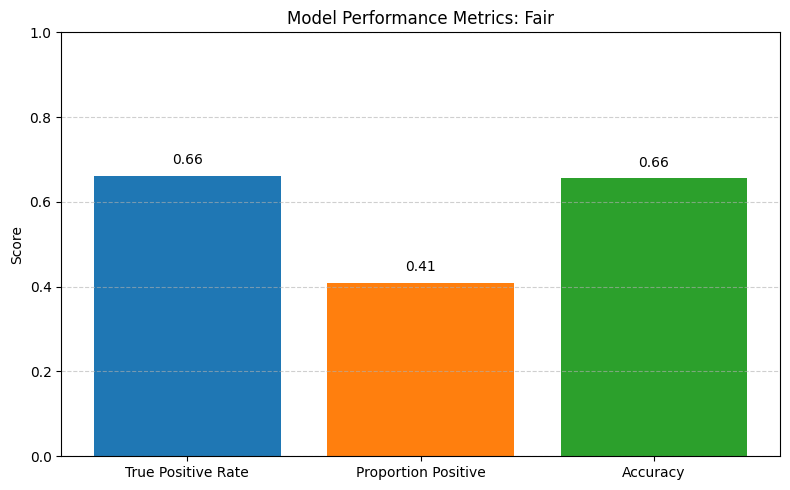}
         \caption{Performance Metrics}
         \label{fig: ...}
     \end{subfigure}
        \caption{Confusion matrices (normalized) for three age groups, and overall performance metrics, after fairness mitigation.}
        \label{fig:confusion_matrix_fair}
\end{figure}

Figure \ref{fig:confusion_matrix_fair} shows the results after applying fairness mitigation, in which separate thresholds are found by group, satisfying FPR under a constraint of balanced accuracy. We observe that normalized false positives (FP) have become much more equal, in fact, FP is now a bit higher for the ``young'' age group. The three normalized confusion matrices look more balanced overall, as can be seen by comparing the color codings. Note that normalized false positives were not directly targeted as the fairness objective; rather, the target was FPR and balanced accuracy. Importantly, the original target of 80\% TPR is no longer satisfied; TPR is now at 66\%. This means that the ``fair'' predictor, capturing only 2/3 of all positive cases, is not as efficient as the original predictor. This kind of ``accuracy-fairness'' tradeoff is to be expected. Note that the overall accuracy is higher after fairness mitigation.

\section{Discussion}
\label{sec:discussion}

\subsection{Contextualization and Stakeholder Perspective}

Let us put the above results in the context of current statistical profiling practices and discuss the ramifications for stakeholders as well as policy lessons of each evaluative dimension.

\textbf{Predictive performance} is one of the most important evaluative dimensions of statistical profiling, it is of high interest to all stakeholder groups. If one takes the predictive performance of the five models evaluated here as an indicator of general predictive performance, then many agencies in the OECD (see Sec. \ref{sec:background}) use suboptimal models: Most agencies that rely on black-box models use RF; the present study found that XGB, the state-of-the-art black-box model, clearly outperforms RF. Turning to interpretable models, most agencies employ a linear model (LR), which is outperformed by EBM. EBM is not currently employed in LTU profiling. Notably, the findings reported here contradict claims (e.g. \citealt{dossc2024}) that black-box models like RF and XGB clearly outperform inherently interpretable models. Rather, the inherently interpretable EBM outperforms RF, and comes close in performance to the best black-box model.

These findings do not directly translate into policy recommendations. For one, as stressed above, predictive performance is only one evaluative dimension. For example, agencies currently employing RF or another black-box model need not necessarily switch to XGB, but could evaluate EBM as an interpretable alternative, which may be superior to RF in terms of predictive performance. What is more, agencies employing LR already have an interpretable model in place, and the cost of disrupting a working and accepted system may be relatively high, because the switch from LR to the predictively superior EBM may pose challenges of interpretation. The same is true for agencies with an operational black-box model.

\textbf{Interpretability} is an important desideratum of statistical profiling for stakeholders. Different stakeholder groups are interested in different aspects of interpretability offered by EBM and LR:

\begin{itemize}

\item[] \textit{Jobseekers} profit from insight into why they have a certain risk prediction. For example, they can inspect the feature function ``insured income'' (Fig. \ref{fig:vers_verd_raw_smooth}) and read off how their income impacts their LTU risk. Both LR and EBM offer the possibility of providing local explanations of single predictions, with the key property that local explanations are faithful to the model, not approximations like explanations of black-box models. On the basis of such explanations (see Fig. \ref{fig:local_explanation} for a fictitious example), jobseekers get the opportunity to adapt their future behavior, but also to contest predictions. Of course, risk prediction can still be incorrect: EBM offers higher predictive performance than LR, but is nevertheless an imperfect, statistical instrument.

\item[] \textit{Caseworkers} profit from the possibility of offering reasons for risk predictions to their clients, for example on the basis of a local explanation, and to develop personalized mitigation strategies on this basis.\footnote{To offer a fictitious example: A caseworker may recommend to lower income expectations to jobseekers with incomes above 6k, which may lower their risk (assuming that insured income is closely related to income expectations); however, if a client has an insured income between 4k and 6k, this may not be advisable.} Interpretable models offer a path to a more participatory use of statistical profiling \citep{delob2021}, which may encourage use of profiling tools and lower resistance to statistical profiling. Both LR and EBM offer a range of different sparsities (Sec. \ref{sec:res_sparsity}). Sparse, smaller models can lower workload and cognitive load of caseworkers. Insight into the risk distributions for single features provided by EBM matters for caseworkers, because this allows them to contextualize single risk predictions. To achieve this, a certain degree of smoothing may be required, also because the justification of single predictions is hard if there are local fluctuations (Sec. \ref{sec:res_smooth}).

\item[] For \textit{policymakers}, the possibility of understanding how certain features influence predictions is an advantage. With interpretable models, they have the possibility to justify the use of statistical profiling. Policymakers can publish the risk model and thereby make it available for public and scientific scrutiny. Using EBM, policymakers can identify, with the use of developers, how current policies influence risk predictions on a granular level.\footnote{For example, both the `age' and the `number of months of previous contributions' have properties that are very likely due to current rules of eligibility for benefits.} Sparsity can matter to policymakers because sparser models require less data, less data collection, are easier to train, and may therefore be more cost effective. 

\item[] \textit{Developers and scientists} can use interpretable models, and in particular information on singular features provided by EBM, to identify problems with data collection. Implausible changes in single EBM feature functions can be discussed with caseworkers, policymakers and domain experts to determine whether they are real effects, due to policies, or point to data issues or problems with the model itself.

\end{itemize}

\textbf{Fairness} mitigation via post-processing was shown to be feasible and also compatible with an interpretable model, for the socially salient attribute of age. Fairness mitigation is a way out of the tradeoff between rule-based profiling and statistical profiling \citep{desie2021}: It achieves fairness, like rule-based profiling, while maintaining adequate predictive performance, like unmitigated statistical profiling -- although a performance penalty is incurred. Fairness is relevant to all stakeholders, but in particular to jobseekers from disadvantages groups. It should be stressed that the post-processing method used here is compatible with any statistical profiling tool, it does not require interpretability. Whether equalizing false positives, as done here, is an adequate fairness intervention, depends on the context of use, and on the utilities or costs imposed on different stakeholders, including jobseekers, but also society at large. Fairness measures, and fairness mitigation, are extremely context dependent. In the context of fairness in machine learning, it is well known that ML models, or statistical modeling, is not value-neutral \citep{delob2021}. Designing statistical profiling tools necessarily involves value-laden, normative choices \citep{vanla2021}. It is extremely important to make choices about intended use and the corresponding choices explicit, because they form the basis for assessments of equity/fairness.

Interpretability itself has at least two dimensions of fairness, which complement the notions of group fairness just discussed. First, it provides a kind of procedural justice \citep{mille2017} by making transparent how individual properties (features) of jobseekers are used to make a prediction. Second, EBM complies with a kind of individual fairness by removing local fluctuations from single feature functions; such local fluctuations, in particular if they are due to overfitting, violate the principle that similar people should be treated similarly. Both of these dimensions are particularly relevant to jobseekers.

\subsection{Limitations and Future Work}

A first limitation of the present work is due to the fact that only administrative data was used. This may limit predictive performance in comparison to, e.g., a combination of administrative and survey data. A thorough survey of features used in different agencies (see e.g. \citealt{desie2019}), and studies combining administrative and survey data, is advisable. Administrative data has the drawback that it is not collected in a controlled manner and with the goal of statistical profiling in mind. For example, the temporal order of data collection is not known for at least some features. This creates the possibility of data leakage, that is, unwanted information about the outcome being contained in the input. Thus, the predictive performance reported here should be viewed as preliminary and with caution. Note that it is expected that the relative predictive performance of the different models would remain the same, at least qualitatively, because data leakage would affect all models, even though there may be quantitative differences in how much different models are affected. To mitigate this issue, extensive data quality control should be conducted, and information about data collection would need to be taken into account.

In previous work on accuracy of statistical profiling in comparison to profiling by caseworkers in Sweden and Switzerland, it was found that statistical profiling is more accurate \citep{desie2019}. It should be investigated how the use of an interpretable statistical profiling tool, with an override option for caseworkers, affects predictive performance and other evaluative dimensions. \citet{vdber2024} have shown that combining predictions on the basis of administrative data with self-assessment by jobseekers and assessment by caseworkers may enhance predictive performance, which speaks in favor of stakeholder involvement.

The work on global interpretability, that is, model sparsity and feature smoothness, presented here is promising, but it could be further improved through stakeholder input, e.g., from data scientists working at unemployment agencies. They are most familiar with data issues that could be identified using single feature functions. Explanations of single predictions, which can be obtained for both EBM and LR, have not been investigated systematically. It should be clarified how such explanations should be displayed to serve caseworkers and jobseekers.

The work on fairness mitigation performed here is preliminary and only establishes the in-principle feasibility of mitigation while preserving interpretability. Group fairness measures other than false positives, other kinds of fairness mitigation, as well as issues with performativity of fairness interventions \citep{zezul2024}, should be investigated, both theoretically and empirically. Ultimately, RCTs should be used to determine the effects of fairness interventions in empirical studies, both in general and with respect to fairness. To deal with the contextuality of fairness in theoretical work, a feasible approach is to work in scenarios, capturing different possible downstream uses of risk predictions; \citet{kern2021} is a step in this direction. Ultimately, statistical profiling tools should be developed with stakeholder involvement, in particular with caseworkers as well as jobseekers, possibly in a participatory design approach \citep{weitz2024}.

\section{Conclusion}
\label{sec:conclusion}

The most important empirical finding of this paper is that interpretable models show predictive performance only slightly worse than the best black-box models. Interpretable models allow stakeholders to understand aspects of statistical profiling that is crucial for them: Insight into general properties of the profiling tool, but also into single predictions, which is particularly relevant to jobseekers and caseworkers. Additionally, interpretable models can be fairness mitigated without loss of interpretability; however, a loss in predictive performance was incurred.

Interpretable profiling is not a one-shot method, but allows all stakeholders to gain actionable insights from profiling tools, to help improving predictions as well as downstream services and activation measures. While this paper is a proof-of-concept, it opens the way to statistical profiling that takes all stakeholders on board. In future research, the current work can be extended by using consolidated and additional (survey) data and stakeholder input to further improve predictive performance, and by considering fairness scenarios. Most importantly, interpretable profiling of long-term unemployment promises the degree of oversight and accountability needed for such high-stakes profiling tools.

\newpage
\addcontentsline{toc}{part}{Appendix}
\begin{center}
{\Large\bfseries Appendix}
\end{center}

\begin{appendices}

\section{Data}

\subsection{Raw Datasets}

The following is a qualitative description of the raw datasets used to construct the clean dataset.The same raw datasets have been used previously in \citep{gasse2022, gasse2023}. For a full list of features of the clean dataset, and which of the raw datasets was used to construct it, see below.

\begin{itemize}

\item \textbf{stes:} Six spreadsheets, data\_stes\_2014-2019, with information on episodes of unemployment of jobseekers (units for which outcomes are available). This includes information on: age; gender; civil status; language skills; industry of last employer; role in last job; desired mobility for job. The episodes of unemployment are indexed by pseudonymized unique identifiers. Jobseekers may have multiple entries, but only one entry per eligibility period is used per jobseeker. The first five years, 2014-2018 were used for training and validation. The set for 2019 was put aside and used for testing.

\item \textbf{asal:} data\_asal contains information from the unemployment benefits accounting system (`Auszahlungssystem Arbeitslosenversicherung'). This includes information on: number of previous months with insurance payments; rate of previous employment; previous income; assignment to invalidity insurance; sickness days; days of prospective benefits. The entries are indexed by month and person, not by unemployment episode.

\item \textbf{beruf:} data\_beruf contains information about the various jobs that a jobseeker looks for (`Beruf'), indexed by unemployment episode. The data contains information such as: assignment of job to Swiss job nomenclature system SBN2000 (e.g. agriculture, skilled labor, sales, ...); experience with job (apprenticeship, expertise, other qualifications); Swiss or foreign certificate; if job was held previously, last, or not at all. Multiple entries may be available for the same unemployment episode, because the data contain information about prospective jobs. 

\item \textbf{arbfo:} data\_arbfo contains information about properties of the job the jobseeker is looking for (`Arbeitsformen'). This includes: working on Sundays and holidays, shift work, night work, and working at home (note on the last: it is unclear how this is related to jobs with home office options). The entries of this table are indexed by episode of unemployment.

\item \textbf{regio:} data\_regio contains the Swiss region in which employment is sought (canton, larger region). It is indexed by unemployment episode.

\item \textbf{outcome:} data\_outco contains the outcome to be predicted, viz. whether or not a jobseeker has received benefits during each of 12 months. The outcome is indexed by month and jobseeker identifier.

\end{itemize}

\subsection{Features Clean Dataset}

\begin{longtable}{p{4.5cm}p{4.5cm}p{1cm}p{1.2cm}p{1cm}p{1.2cm}}
\hline
Variable Name & Semantics & Source & Range & Num. & Dummy \\
\hline
\endfirsthead

\hline
Variable Name & Semantics & Source & Range & Num. & Dummy \\
\hline
\endhead

\hline
\multicolumn{6}{r}{Continued on next page} \\
\hline
\endfoot

\hline
\endlastfoot

alter & age & stes & int. & Y & N \\
anz\_b\_ausgeuebt & \# jobs held & beruf & int. & Y & N \\
anz\_b\_erf\_0 & \# jobs 0 exp. & beruf & int. & Y & N \\
anz\_b\_erf\_1bis3j & \# jobs 1-3 years exp. & beruf & int. & Y & N \\
anz\_b\_erf\_1j & \# jobs 1-less years exp. & beruf & int. & Y & N \\
anz\_b\_erf\_3j & \# jobs 3-plus years exp. & beruf & int. & Y & N \\
anz\_b\_ges\_ausl\_abs & \# jobs s. \& foreign cert. & beruf & int. & Y & N \\
anz\_b\_ges\_exp\_lehre & \# jobs s. \& appr. qual. & beruf & int. & Y & N \\
anz\_b\_ges\_exp\_quali & \# jobs s. \& expert qual. & beruf & int. & Y & N \\
anz\_b\_ges\_gelernt & \# jobs searched \& learned & beruf & int. & Y & N \\
anz\_b\_ges\_inl\_abs & \# jobs Swiss cert. & beruf & int. & Y & N \\
anz\_b\_gesucht & \# jobs searched & beruf & int. & Y & N \\
anz\_b\_gesucht\_zuletzt & \# jobs searched \& held last & beruf & int. & Y & N \\
anz\_b\_mit\_erf\_such & \# jobs searched \& exp. & beruf & int. & Y & N \\
anz\_b\_n\_ausgeuebt & \# jobs not held & beruf & int. & Y & N \\
ar\_mue\_bin & language Ara. oral & stes & 2 ivs.  & N & Y \\
aufenthalt\_bins & residence status & stes & binary  & N & N \\
ausbildung\_bins & education level & stes & 4 ivs.  & N & Y \\
ausweis\_b\_b1\_be & driving permit & asal & binary  & N & N \\
b\_ges\_ausg\_sek & \# job s. \& h. \& sec. edu. & beruf & int. & Y & N \\
b\_ges\_ausg\_ter & \# job s. \& h. \& tert. edu. & beruf & int. & Y & N \\
beitragsmonate\_vor\_rf & \# months w/ prev. contributions & asal & int. & Y & N \\
berufskl\_1 & \# jobs in class 1 (snb2000) & beruf & int. & Y & N \\
berufskl\_2 & \# jobs in class 2 `` '' & beruf & int. & Y & N \\
berufskl\_3 & \# jobs in class 3 `` '' & beruf & int. & Y & N \\
berufskl\_4 & \# jobs in class 4 `` '' & beruf & int. & Y & N \\
berufskl\_5 & \# jobs in class 5 `` '' & beruf & int. & Y & N \\
berufskl\_6 & \# jobs in class 6 `` '' & beruf & int. & Y & N \\
berufskl\_7 & \# jobs in class 7 `` '' & beruf & int. & Y & N \\
berufskl\_8 & \# jobs in class 8 `` '' & beruf & int. & Y & N \\
berufskl\_9 & \# jobs in class 9 `` '' & beruf & int. & Y & N \\
beschaeftigungsgrad\_vorher & rate of prev. employment & asal & int. & Y & N \\
cat\_geschlecht & gender & stes & binary  & N & N \\
ch\_mue\_bin & language Sw.-Ger. oral & stes & 2 ivs.  & N & Y \\
cod\_esa & job situation at registr. & stes & 8 int.  & N & Y \\
cod\_mobilitaet & desired mobility for job & stes & 5 int.  & N & Y \\
cod\_zivilstand & civil status & stes & 4 int.  & N & Y \\
code\_arbeitsform & mode of work & arbfo & 6 int.  & N & Y \\
code\_funktion & role last job & stes & 9 int.  & N & Y \\
de\_sch\_bin & language Ger. written & stes & 2 ivs.  & N & Y \\
en\_mue\_bin & language  Eng. oral & stes & 2 ivs.  & N & Y \\
en\_sch\_bin & language  Eng. written & stes & 2 ivs.  & N & Y \\
es\_mue\_bin & language  Spa. oral & stes & 2 ivs.  & N & Y \\
fr\_mue\_bin & language Fre. oral & stes & 2 ivs.  & N & Y \\
fr\_sch\_bin & language  Fre. written & stes & 2 ivs.  & N & Y \\
it\_mue\_bin &  language  Ita. oral & stes & 2 ivs.  & N & Y \\
it\_sch\_bin & language Ita. written & stes & 2 ivs.  & N & Y \\
iv\_code\_bins & invalidity insurance & asal & 2 ivs.  & N & Y \\
krankentagg\_bins & sickness days & asal & 2 ivs.  & N & Y \\
monat\_kal & calendar month & stes & 12 int. & N & Y \\
noga\_bins & industry of last employment & stes & 21 ivs.  & N & Y \\
pauschale\_bins & reason for lump sum & asal & 2 ivs.  & N & Y \\
prozentsatz\_bins & \% of income benefits & asal & 2 ivs. & N & Y \\
suchreg\_gross & search region  & regio & 6 int.  & N & Y \\
taggeld\_anspr\_bins & days of insurance claims & asal & 7 ivs. & N & Y \\
vermittlungsgrad\_asal & desired rate of employment & asal & int. & Y & N \\
vers\_verdienst & insured income & asal & int. & Y & N \\
\end{longtable}

Abbreviations: appr.: apprenticeship; cert.: certification; empl.: employment; exp.: experience; h.: held; int.: integer; ins.: insurance; ivs.: intervalls; lang.: language skills; mob.: mobility; qual.: qualification; registr.: registration; s.: searched; sbn2000: Swiss job nomenclature 2000\footnote{Publicly available via Swiss Federal Statistical Office.}; sec.: secondary, Sw.Ger.: Swiss-German; tert.: tertiary.

For a description of `Source' see raw data above.

\section{Hyperparameters}

\begin{table}[H]
\centering
\begin{tabular}{|c|c|}
\hline
Model & Hyperparameters \\
\hline
RF & 'n\_estimators': 500, 'max\_depth': 10, 'max\_features': 50\\
\hline
GB & 'n\_estimators': 500, 'learning\_rate': 0.1, 'max\_depth': 6\\
\hline
XGB &  'n\_estimators': 500 , 'learning\_rate': 0.1, 'max\_depth': 6, \\
& 'reg\_lambda': 5,  'objective': 'binary:logistic' \\ 
\hline
EBM & 'interactions': 30, 'outer\_bags': 9, 'learning\_rate': 0.0014, \\ 
& 'min\_samples\_leaf': 2, 'max\_leaves': 3 \\
\hline
LR & 'l\_1': 0.1 \\
\hline
\end{tabular}
\caption{Hyperparameter Settings}
\label{tab:hyperparameters}
\end{table}

For XGB and EBM, hyperparameter search was performed using the Optuna package \citep{akiba2019}, a state-of-the-art hyperparameter tuning framework. We thus get a ``fair'' comparison of the best black-box and the best interpretable model. For XGB, 50 trials in Optuna over all folds and available hyperparameters did not yield a substantive gain in AUC, with a best average AUC of 0.7743; it was therefore decided to retain the original parameters. For EBM, it was found that the main hyperparameter affecting performance was the number of interactions, with a higher number of interactions leading to better performance. It was decided to set the number of interactions to a (low) value by hand, because the main goal was to obtain an interpretable version of EBM, which includes limiting the number of features and interactions. Note that hyperparameter tuning is relatively cheap for XGB, but expensive for EBM, because the latter model is much slower to fit. Hyperparameters were not tuned for RF and GB; values were taken from \citet{gasse2022}, who optimized these models on the same data, but in a different context.

\section{Additional Results}

\subsection{Predictive Performance}

\begin{table}[ht]
\centering
\begin{tabular}{|c|c|c|c|c|c|}
\hline
Year & LR & RF & {\bf XGB} & GB & EBM \\
\hline
2015 & 0.7523 & 0.7565 & 0.7750 & 0.7684 & 0.7681 \\
2016 & 0.7496 & 0.7560 & 0.7739 & 0.7721 & 0.7674 \\
2017 & 0.7482 & 0.7518 & 0.7706 & 0.7686 & 0.7630 \\
2018 & 0.7443 & 0.7490 & 0.7697 & 0.7670 & 0.7607 \\
$2019^*$ & 0.7589 & 0.7435 & 0.7619 & 0.7504 & 0.7593 \\
\hline
\end{tabular}
\caption{Results (AUC) by validate year (2015-2018) and test year (2019). Results are displayed graphically in Fig. \ref{fig:scores_train_test}.}
\label{tab:results total auc}
\end{table}

\subsection{Interpretability: Sparsity}

\begin{figure}[H] 
\centering
\includegraphics[width=0.8\textwidth]{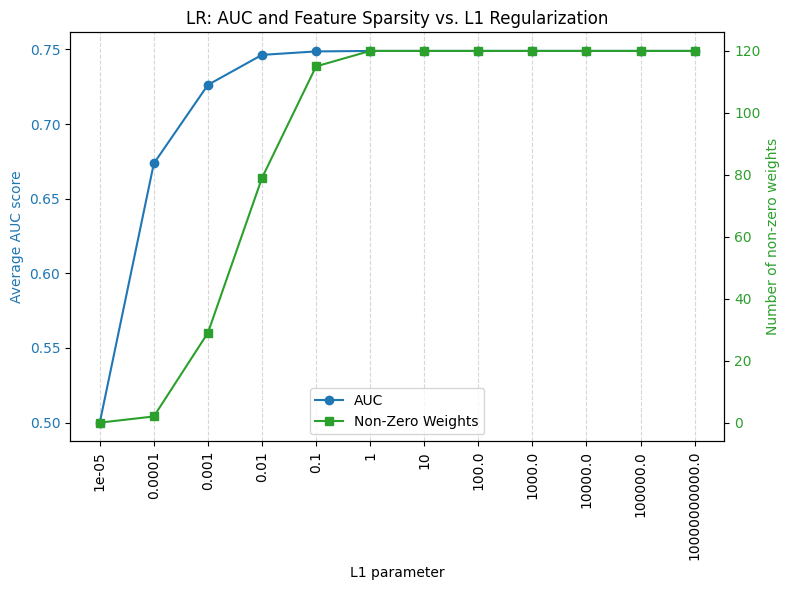}
\caption{Effect of $L_1$ regularization on number of non-zero features and predictive performance of LR. Note that an AUC of 0.5 corresponds to random predictions (uninformative model).}
\label{fig:lr_auc_sparsity}
\end{figure}

\begin{figure}[H] 
\centering
\includegraphics[width=0.9\textwidth]{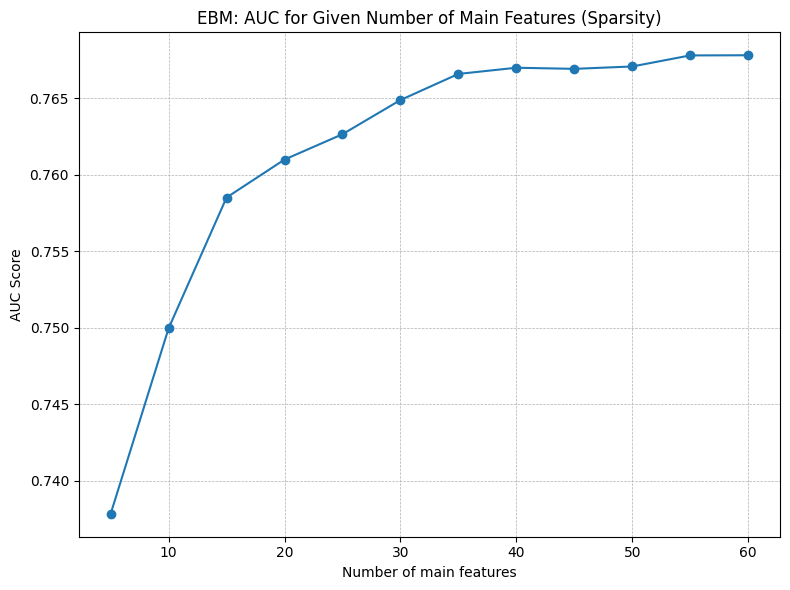}
\caption{Performance Score (AUC) for EBMs obtained from backward selection, retaining $n$ most important main features in steps of 5. For each $n$, a fraction of $0.5$ interactions of the total number of main effects were added; the rightmost model, with 57 main features and 30 interactions, is the full EBM.}
\label{fig:ebm_auc_sparsity_5}
\end{figure}

\begin{figure}[H] 
\centering
\includegraphics[width=\textwidth]{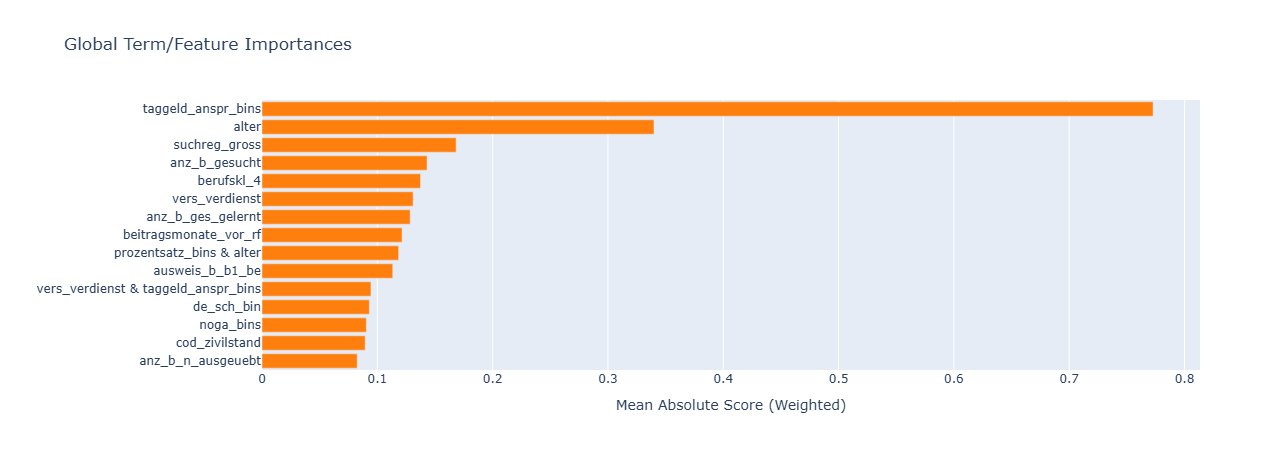}
\caption{Global explanation (feature importance plot) generated by EBM interface. It shows the (average) feature importance of the 15 most important features for fold 1 (trained on data from 2014). Features with `\&' are interactions.}
\label{fig:ebm_global_explanation}
\end{figure}

\subsection{Interpretability: Smoothing}

\begin{figure}[H] 
\centering
\includegraphics[width=0.9\textwidth]{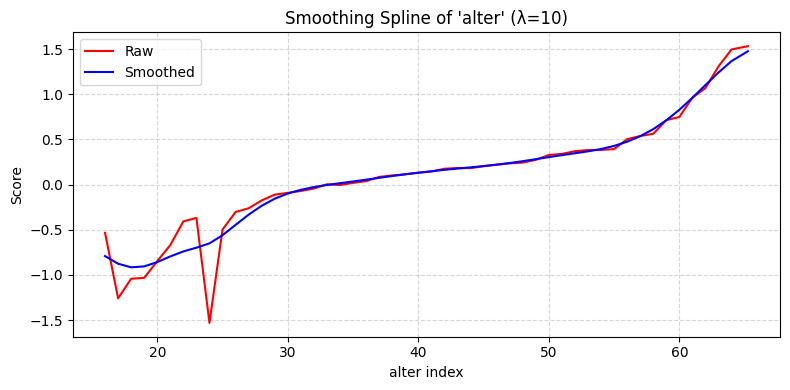}
\caption{Numerical feature `age' (alter), in raw version (off-the-shelf EBM) and after applying smoothing spline with smoothing parameter $\lambda$; age in years.}
\label{fig:alter_raw_smooth}
\end{figure}

\begin{figure}[H] 
\centering
\includegraphics[width=0.9\textwidth]{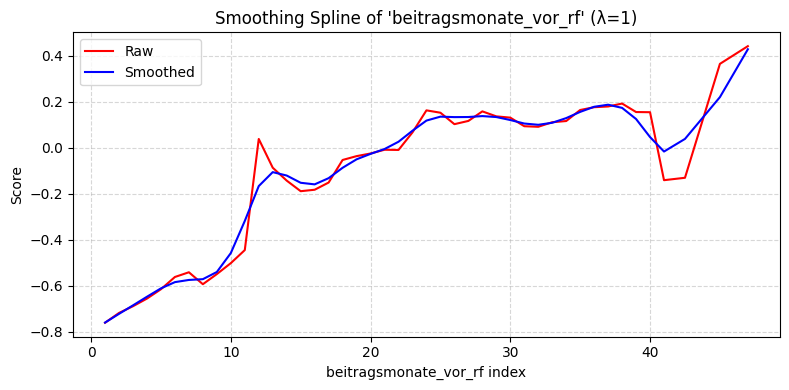}
\caption{Numerical feature `number of months of previous contributions' (beitragsmonate\_vor\_rf), in raw version (off-the-shelf EBM) and after applying smoothing spline with smoothing parameter $\lambda$.}
\label{fig:beitragsmonate_raw_smooth}
\end{figure}

\begin{figure}[H] 
\centering
\includegraphics[width=0.9\textwidth]{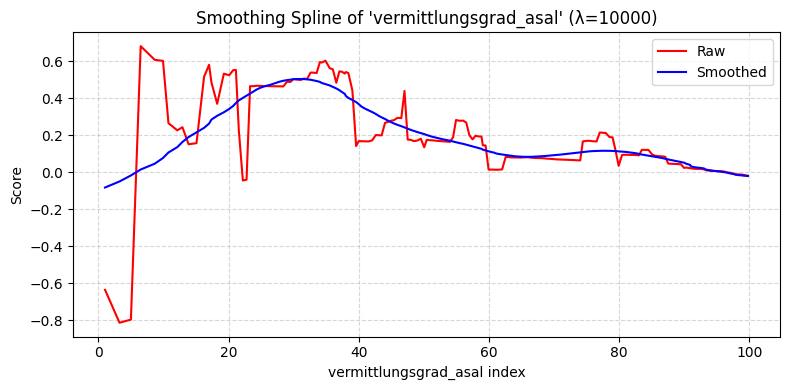}
\caption{Numerical feature `rate of desired employment' (vermittlungsgrad\_asal), in raw version (off-the-shelf EBM) and after applying smoothing spline with smoothing parameter $\lambda$.}
\label{fig:vermittlungsgrad_raw_smooth}
\end{figure}

\subsection{Interpretability: Local Explanation}

\begin{figure}[H] 
\centering
\includegraphics[width=\textwidth]{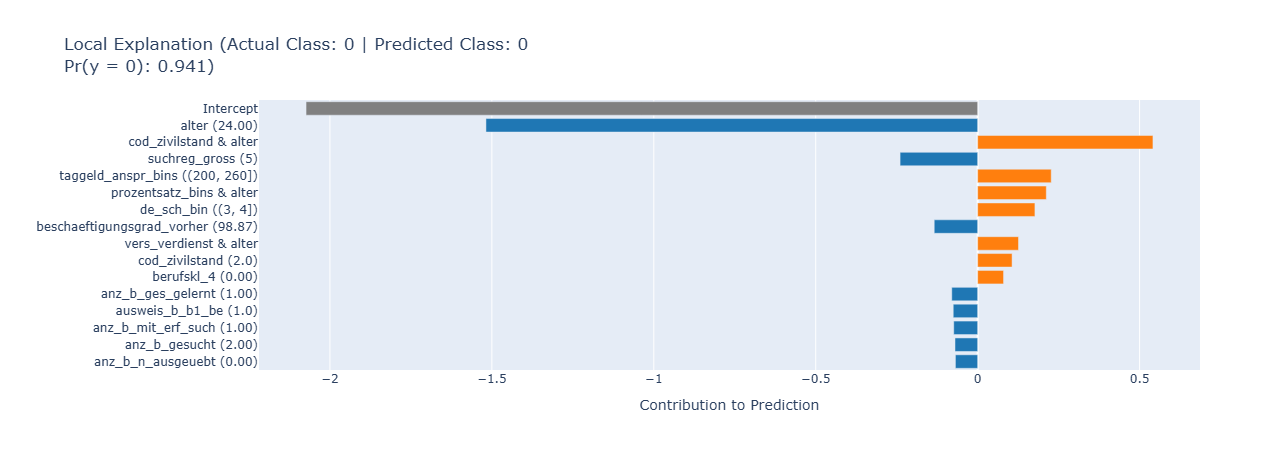}
\caption{Example of a local explanation generated by EBM interface for a full EBM. Fictitious example with random features and outcome. Contributions of single features and interactions (two features connected by `\&') to overall risk score are ordered in descending absolute importance; blue features contribute negatively (lower score), orange features contribute positively (higher score). Only 15 most important features and intercept (fixed) are listed.}
\label{fig:local_explanation}
\end{figure}

\subsection{Fairness}

\begin{figure}[H] 
\centering
\includegraphics[width=0.8\textwidth]{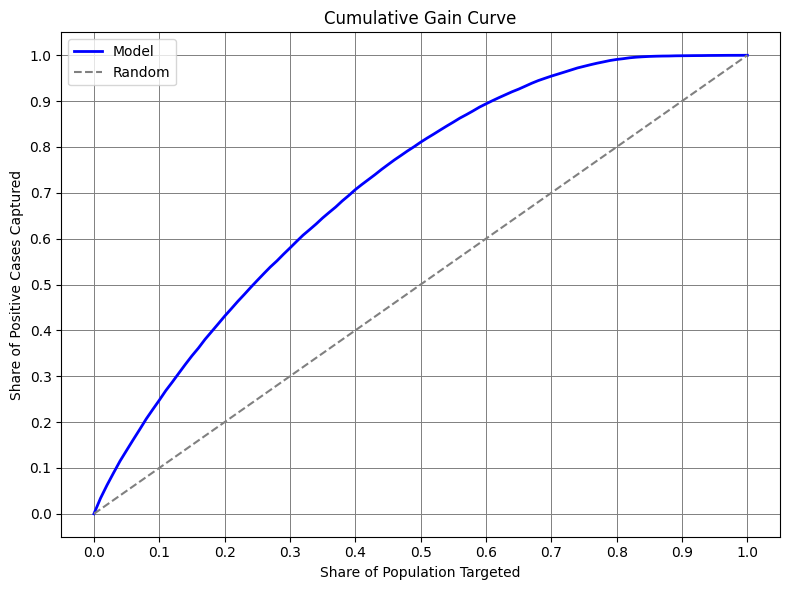}
\caption{Cumulative gains curve for EBM-30 model. This curve can serve as a basis for the choice of threshold. In the hypothetical fairness scenario, a level of positive cases to be captured (80\%) was chosen; one can the read off the cumulative gains curve that in this case, one has to target approximately 49\% of the total population.}
\label{fig:lr_auc_sparsity}
\end{figure}

\section{Reproducibility}

\subsection{Data Availability}

The data used in this study is not publicly available due to privacy restrictions. However, the data can be obtained from the Swiss State Secretariat for Economic Affairs (SECO), section ``Arbeitsmarkt / Arbeitslosenversicherung'', under certain conditions and with data privacy measures in place. The datasets used in this work can thus be obtained from a third party for replications and extensions of the present work. Feel free to contact the author for more information about how to obtain data.

\subsection{Code}

Full code to reproduce all results is available on GitHub: \href{https://github.com/timraez/interpretable-fair-profiling}{github.com/timraez/interpretable-fair-profiling}.

\end{appendices}

\end{document}